\documentclass[runningheads]{llncs}
\usepackage[T1]{fontenc}
\usepackage{amssymb}
\usepackage{amsmath}
\usepackage{subcaption}
\usepackage{booktabs}
\usepackage{svg}
\usepackage{multirow}
\usepackage{graphicx}
\usepackage{url}
\usepackage[section]{placeins}
\begin{document}
\title{ViTs for action classification in videos: An approach to
risky tackle detection in American football practice videos}
\titlerunning{Risky Tackle Detection in American Football}
%
\author{Syed Ahsan Masud Zaidi\inst{1}\and
William Hsu\inst{1} \and
Scott Dietrich\inst{2}}
%

%
\institute{Kansas State University, Manhattan , KS ,66506, USA
\\
\and
Albright College, Athletics/Athletic training, Reading, PA,1962, USA\\
}
\maketitle              
\begin{abstract}
Early identification of hazardous actions in contact sports enables timely intervention and improves player safety. We present a method for detecting \emph{risky} tackles in American football practice videos and introduce a substantially expanded dataset for this task. Our work contains 733 single-athlete-dummy tackle clips, each temporally localized around first point contact and labeled with a strike zone component of the standardized Assessment for Tackling Technique (SATT-3), extending prior work that reported 178 annotated videos.

Using a Vision transformer-based model with imbalance-aware training, we obtain \textbf{risky recall} of \textbf{0.67} and \textbf{Risky F1} of \textbf{0.59} under cross-validation. Relative to the previous baseline in a smaller subset (risky recall of \(0.58 \); Risky $F1$ \(0.56\) ), our approach improves risky recall by more than \textbf{8\%}  points on a much larger dataset. These results indicate that the vision transformer-based video analysis, coupled with careful handling of class imbalance, can reliably detect rare but safety-critical tackling patterns, offering a practical pathway toward coach-centered injury prevention tools.

\end{abstract}
\section{Introduction}
Player safety in American football hinges on the early identification of hazardous tackling mechanics, particularly during habit formation when coaching interventions can effectively modify behavior. Coaches and trainers increasingly utilize practice videos for this purpose; however, the cues distinguishing risky tackles from safe ones are subtle, transient, and often overlooked at full speed. Manual review fails to scale across hundreds of sessions and suffers from inconsistency, particularly given the rarity of critical events. A reliable automated video-based risk detector could prioritize reviews, standardize feedback, and ultimately mitigate the risk of head and neck injuries

We formalize this task as a binary classification of tackles as risky or safe in practice videos, focusing on a short interval near the point of contact and evaluated using a standardized tackling rubric (SATT) \cite{Dietrich_SATT1_PlayerControl,Dietrich2019SATTReliability,Dietrich2020SATTValidation,Nafi2022RiskyTackle3D}.
 We evaluated tackles using the SATT rubric, specifically the SATT3 (Strike Zone, SZ)  criterion at the point of contact, which emphasizes keeping the head clear of the target and initiating contact with the front of the chest/top of the shoulder. Videos were labeled by domain experts on a 0–3 scale (Did-Not-Occur, Poor, Average, Excellent); we map scores $\leq$ 1 to ‘risky’ and scores $\geq$ 2 to ‘safe’.  A score of \textbf{0 (did not occur)} indicates head-first/crown-exposed contact (spearing) with little or no shoulder involvement and poor arm preparation. Score \textbf{1 (poor)} indicates the head still makes contact and the tackler does not initiate with the chest or shoulder, with arms not engaged; \textbf{2 (average)} indicates the head stays clear and to the side, with primary contact delivered by the \emph{top of the shoulder} and the arms cocked and ready to wrap; \textbf{3 (excellent)} indicates the head and eyes remain clear of the target throughout, with primary contact delivered by the front of the chest (or a proper shoulder strike) and active arm engagement to secure the tackle.

Some key practical challenges in this study is dealing with class imbalance (with safe tackles predominating), limited labeled data, and significant nuisance variations across sessions (e.g., camera angles, lighting conditions, and athlete physiques). Consequently, there is a need for an effective solution that may prioritize recall for the risky class while maintaining overall reliability, ensuring reproducible labeling and precise temporal localization, and functioning robustly on authentic practice footage rather than curated game highlights.

Capturing risky tackles presents inherent challenges. They are rare by nature, ethically challenging to simulate, fleeting at the point of initial contact, and frequently obscured by players or gear. Consequently, practice videos exhibit a natural imbalance of approximately \textbf{$65\%$} safe and \textbf{35\%} risky with annotations demanding expert evaluation via the SATT rubric, thereby constraining dataset scalability. This imbalance yields a misleading baseline, wherein a trivial classifier predicting \emph{safe} for all instances achieves \textbf{65\%} accuracy yet entirely overlooks the critical events. In addition to class imbalance and annotation scarcity, extraneous variations (e.g., camera positioning, illumination, and inter-athlete differences) and the ephemeral, nuanced visual indicators of dangerous techniques make the problem more challenging. 

The key contributions of this study are threefold. First, we expand the empirical basis for risky-tackle analysis by curating a substantially larger corpus of real-world practice footage, annotated with SATT labels and precise first-contact timestamps. Second, we address severe class imbalance via four types of augmentations that increase the prevalence of risky events while preserving a realistic distribution in validation. Thirdly, we trained a class imbalance-aware vision transformer to further mitigate class imbalance and reliably detect risky tackles in training videos.

\section{Background Work}

Video understanding for sports has advanced rapidly, spanning action recognition, temporal spotting, player tracking, and coach-facing analytics \cite{wu2022surveyvideoactionrecognition,xu2025actionspottingpreciseevent}. Recent efforts also move beyond closed-set classification toward question-driven and domain specialized reasoning about actions, reflecting the practical needs of analysts and coaches \cite{salehi2024actionatlasvideoqabenchmarkdomainspecialized}. Across these settings, practice footage poses distinct, limited constraints on camera control, variable viewpoints, and brief, safety critical events that differ from broadcast highlights and demand task-specific benchmarks and evaluation.

A persistent challenge in sports video understanding is robustness to occlusion and background clutter, as interactions frequently mask key body parts at decisive instants \cite{NEURIPS2023_Grover2023,modi2024occlusionsvideoactiondetection}. Closely related is the task of \emph{action spotting} and precise event detection, which emphasizes localizing short, semantically meaningful moments (e.g., ball contact, collisions) under temporal sparsity and ambiguity \cite{xu2025actionspottingpreciseevent,ASTRA_2024}. Recent video backbones spanning convolutional architectures and attention based models seek stronger spatiotemporal representations and long-range context aggregation \cite{SlowfastVideo2019,TimeSformer2021,arnab2021vivit,LIU2025125642,mvitv2}.

Within American football, prior work has focused on formation analysis and automated annotation for strategy, as well as tracking/classification from game film \cite{electronics12030726,Newman2021}. Parallel threads in the broader literature study egocentric cues, prototype based representations, and human-centric motion quality, highlighting the community’s interest in evaluating how an action is performed, not just what it is \cite{wang_2021,LIU2025125642,huang2025hmorelearninghumancentricmotion}.The real world deployment contexts favor robust and resource-efficient monitoring pipelines \cite{nayab2025advancing} and accurate decision-making benefits from stabilizing measurements via mechanisms analogous to disturbance rejection in closed-loop tracking systems \cite{zaidi2019positioncontrol_fso,khan2018olympus} .
The initial tackle training dataset for practice videos for risky tackle detection was introduced by Nafi et al. \cite{Nafi2022RiskyTackle3D}. They implemented a C3D-based model to identify the risky tackles. Their implementation was based on the temporal and spatial localization of training videos to be classified using C3D for risky tackle detection.
Although some work has been done previously, the safety-oriented assessment of tackles in practice settings remains widely underexplored: datasets are small, risky events are rare by design, and care must be taken when training a model with imbalanced data. This gap motivates a focus on standardized labeling of tackle quality and on benchmarks that reflect the operational needs of injury-prevention workflows.

\section{Methodology}

\subsection{Dataset Preparation}
In this work, we expanded the previously published dataset of 178 practice videos \cite{Nafi2022RiskyTackle3D} by introducing 555 new tackle training videos to the dataset. Each video represents a training practice tackle where a player runs and makes contact with the dummy. The diversity in videos helps create a generalized data set that consists of video samples of a variety of background scenes, multi-dressed, geared, and non-gearred players. The summary of the enhanced dataset is shown in Table \ref{tab:dataset_summary}.
\begin{table}[htbp]
\centering
\caption{Tackle Video Dataset distribution by class }
\label{tab:dataset_summary}
\begin{tabular}{lrr}
\toprule
Class & Count & Percentage (\%)\\
\midrule
Safe  & 474 & 64.7\\
Risky  & 259 & 35.3\\
\midrule
\textbf{Total} & \textbf{733} & \textbf{100.0}\\
\bottomrule
\end{tabular}
\end{table}


Videos were recorded under diverse lighting and background conditions, with players wearing varied attire to promote generalization in practical training scenarios. Clip durations ranged from $200$ to $1500$ frames at $30 fps$. Each video was manually annotated using the SATT rubric, deriving labels from \emph{SATT3 (Strike Zone, SZ)} at the point of contact (head clear of target; chest/top-of-shoulder contact). We mapped the 0--3 scores to binary classes: scores $\leq 1$ as \emph{risky} and scores $\geq 2$ as \emph{safe}.

\subsection{Augmentation Implementation and Class Balancing}
To mitigate the class imbalance between risky and safe tackles, we augmented samples from the minority (\emph{risky}) class and reintegrated them into the training set. We also augmented a subset of the majority (\emph{safe}) class to reduce overfitting to augmentation artifacts, encouraging the model to rely on biomechanical cues rather than augmentation-specific patterns. We used four augmentation types (Table~\ref{tab:augmentations}) with discrete levels: 2 for Noise and 3 each for Brightness, Rotation, and Flip, yielding $2 \times 3 \times 3 \times 3 = 54$ configurations. Evaluating all 54 combinations exhaustively is computationally expensive.



\begin{figure}[htbp]
  \centering
  \includegraphics[width=0.8\textwidth]{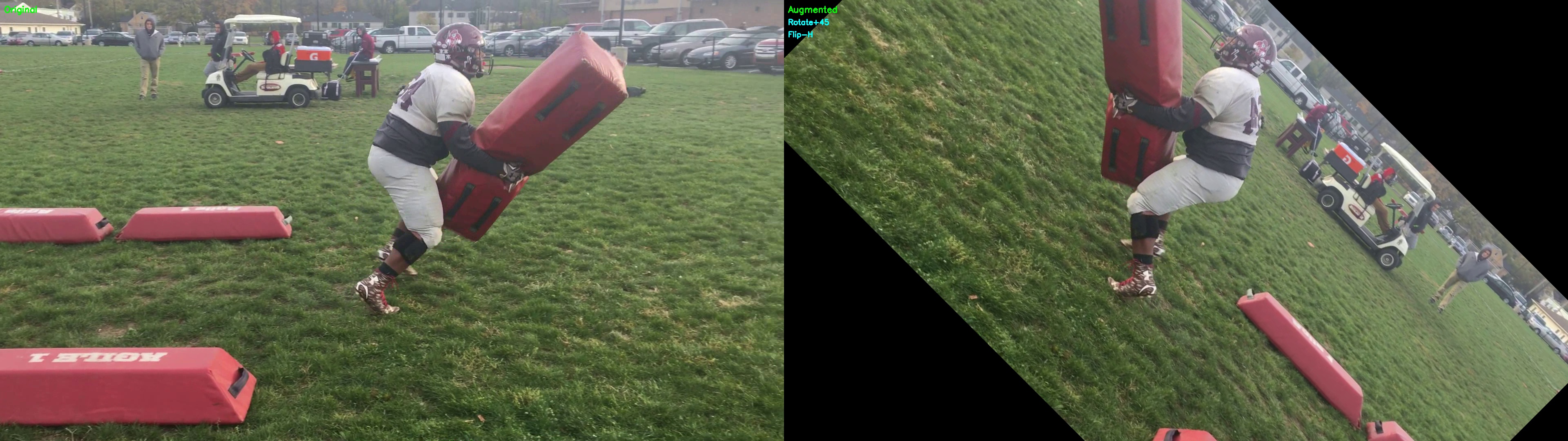}
  \caption{Sample original (left) vs.\ augmented (right) frames for \textbf{Run\(_8\)}.}
  \label{fig:Augmentation_examples}
\end{figure}

\begin{table}[htbp]
\centering
\caption{Augmentations and their level options.}
\label{tab:augmentations}
\begin{tabular}{l l c}
\toprule
\textbf{Augmentation} & \textbf{Levels} & \textbf{Count} \\
\midrule
Noise       & None; Noise                         & 2 \\
Brightness  & Same; Increase; Decrease            & 3 \\
Rotate      & None; Left; Right                   & 3 \\
Flip        & None; Horizontal; Vertical          & 3 \\
\midrule
\multicolumn{2}{r}{\textbf{Total combinations}} & \textbf{54} \\
\bottomrule
\end{tabular}
\end{table}

To efficiently explore this combinatorial space, we employed Taguchi's orthogonal arrays \cite{phadke1989quality}, a design of experiments methodology that enables systematic investigation of parameter interactions with minimal experimental runs. Using the $L_{18}$ orthogonal array design, we reduced the combinatorial space from 54 to just 18 configurations, significantly decreasing computational requirements while maintaining statistical rigor. Details of the $L_{18}$ experiment runs are provided in table \ref{tab:l18_aug_schedule}

Augmentation is applied selectively to achieve two objectives: (1) balance the training class distribution to prevent majority-class bias, and (2) systematically explore the augmentation parameter space to identify optimal strategies. The Taguchi $L_{18}$ orthogonal array defines 18 augmentation configurations plus one baseline (no augmentation), yielding 19 distinct experimental runs. Each configuration specifies combinations of four augmentation factors: \textbf{ Gaussian noise injection, brightness adjustment ($50\%$ in HSV V-channel), rotation ($45\deg$), and flipping (horizontal/vertical)}.

\begin{table}[htbp]
\centering
\caption{Taguchi $L_{18}$ schedule for augmentation factors.}
\label{tab:l18_aug_schedule}
\begin{tabular}{@{}cllll@{}}
\toprule
\textbf{Run  } & \textbf{Noise} & \textbf{Brightness} & \textbf{Rotate} & \textbf{Flip} \\
\midrule
R1  & None      & Increase & Left  & Horizontal \\
R2  & None      & Increase & Right & Vertical   \\
R3  & None      & Increase & None  & None       \\
R4  & None      & Decrease & Left  & Vertical   \\
R5  & None      & Decrease & Right & None       \\
R6  & None      & Decrease & None  & Horizontal \\
R7  & None      & None     & Left  & None       \\
R8  & None      & None     & Right & Horizontal \\
R9  & None      & None     & None  & Vertical   \\
R10 & Add noise & Increase & Left  & None       \\
R11 & Add noise & Increase & Right & Horizontal \\
R12 & Add noise & Increase & None  & Vertical   \\
R13 & Add noise & Decrease & Left  & Horizontal \\
R14 & Add noise & Decrease & Right & Vertical   \\
R15 & Add noise & Decrease & None  & None       \\
R16 & Add noise & None     & Left  & Vertical   \\
R17 & Add noise & None     & Right & None       \\
R18 & Add noise & None     & None  & Horizontal \\
\bottomrule
\end{tabular}
\end{table}

Critically, augmentation is applied \textbf{after stratified fold splitting} and \textbf{only to the training data}. For each fold, the augmentation pipeline generates $\sim$171--173 additional videos by applying Taguchi-specified transformations to selected originals. Augmentation primarily targets the minority (\emph{risky}) class to achieve an approximately 50:50 training balance, while validation sets remain unaugmented to preserve natural class distributions for unbiased evaluation. Beyond the 18 Taguchi runs, we included two baselines: \textbf{Run\(_0\)} uses oversampling (duplicate risky videos) without transformations to test whether class rebalancing alone improves learning, and \textbf{Run\(_\mathrm{orig}\)} uses the original unbalanced dataset to quantify the benefit of augmentation. A visual example for $run_18$ is shown in figure \ref{fig:Augmentation_examples}

\subsection{Temporal Localization in videos}
\begin{figure}[!htbp]
  \centering
  \includegraphics[width=0.9\textwidth]{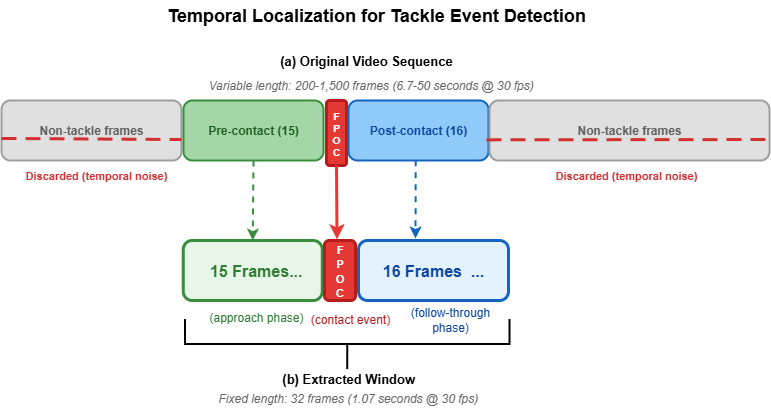}
  \caption{Temporal localization: Extracting tackle event from raw videos.}
  \label{fig:temporal_localization}
\end{figure}

A key challenge is the tackle event detection in the variable duration of the videos, with the tackle occurring at arbitrary times. We therefore perform temporal localization to isolate the event and reduce non-tackle noise. Specifically, we manually mark the first point of contact (FPOC) between the player and dummy, then extract a fixed 32-frame clip (15 frames before and 16 frames after the FPOC) as the standardized model input (Fig.~\ref{fig:temporal_localization}).

\begin{figure}
    \centering
    \includegraphics[width=1\linewidth]{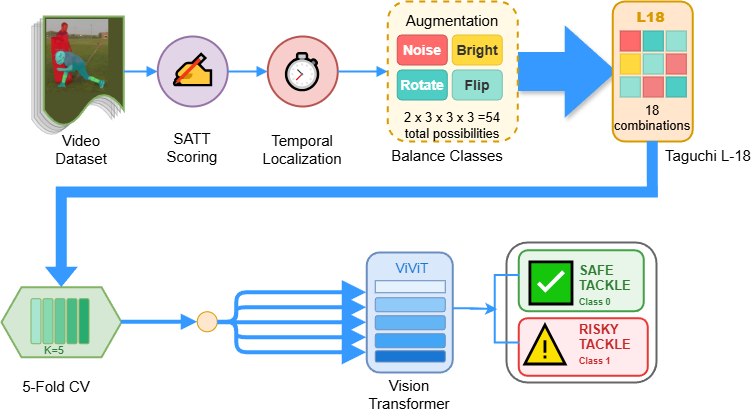}
    \caption{Pipeline overview: first-contact localization with fixed-window trimming, Taguchi $L_{18}$-guided augmentation, stratified 5-fold cross-validation, and ViViT training for Risky Tackle Detection}
    \label{fig:pipeline}
\end{figure}

\subsection{Stratified 5-Fold Cross-Validation with Post-Split Augmentation}

The dataset is class-imbalanced (733 videos: $\sim$65\% safe, $\sim$35\% risky). We therefore use stratified 5-fold cross-validation so each validation fold matches the global class ratio. Folds are created with \texttt{StratifiedKFold(n\_splits=5, shuffle=True, random\_state=42)}, yielding $|S_k|\in\{146,147\}$ with roughly 51--52 risky and 94--95 safe videos per fold. For each fold $k$, we train on $\mathcal{D}\setminus S_k$, evaluate on $S_k$, and report mean$\pm$SD over the five folds.

To reduce majority-class bias while preserving realistic evaluation, we apply \emph{post-split} augmentation: after splitting, we augment only the training portion by synthesizing risky clips to reach an approximately 50:50 class balance, while keeping validation folds unchanged. Per fold, we generate
\[
N_{\text{aug}} = N_{\text{safe}}^{\text{train}} - N_{\text{risky}}^{\text{train}} \approx 171\text{--}173
\]
augmented risky clips using Taguchi-designed transformations, producing $\sim$757--760 training clips. Leakage is prevented by (i) never augmenting validation videos and (ii) ensuring each original appears in validation exactly once.

We train one ViViT model per fold for each setting (19 Taguchi configurations plus one unbalanced baseline), totaling $5\times20 =100$ runs. Models are initialized from Kinetics-400 pretraining, trained with early stopping (macro-F1), and evaluated on the imbalanced validation folds. Metrics are reported as mean$\pm$SD, with risky recall as the primary endpoint.
Across folds, training sets contain $586-587$ originals plus $171-173$ augmented risky clips $(758-760$ total) , yielding $\approx50:50 $ balance; validation sets remain unaugmented with $146-147$ videos and $34.9-35.6\% $ risky.


\subsection{Video Vision Transformer Architecture}

We employ the ViViT-B 16$\times$2 architecture~\cite{arnab2021vivit}, pretrained on Kinetics-400~\cite{kay2017kineticshumanactionvideo} . Kinetics-400 comprises $\sim$400 human action classes with over 300,000 videos, providing rich spatiotemporal representations that transfer effectively to downstream video understanding tasks.

The model designation indicates: base size (hidden dimension $D=768$, 12 transformer layers), 16$\times$16 pixel spatial patches, and tubelet size of 2 frames (temporal patches span 2 consecutive frames). For our 32-frame $\times$ 224$\times$224 inputs, this yields $\frac{32}{2} \times \frac{224}{16} \times \frac{224}{16} = 3{,}136$ spatiotemporal tokens plus one classification token.

\textbf{Transfer learning rationale:} Training ViViT from scratch on Kinetics 400 requires weeks of multi-GPU computation. Pretrained weights provide three advantages: (1) computational efficiency, reducing training time from weeks to hours; (2) improved generalization—pretrained spatiotemporal features (motion patterns, body dynamics, temporal relationships) transfer to tackle analysis; (3) effective learning with limited data, critical given our 733-video dataset, modest compared to Kinetics-400's 300K+ videos.

We employ full fine-tuning: all encoder parameters (initialized from Kinetics-400) and the classification head (randomly initialized, adapted from 400 to 2 classes) are updated during training. This balances leveraging pretrained representations while adapting to tackle-specific spatiotemporal patterns.

\subsubsection{Architecture Overview}

ViViT processes video input $\mathbf{V} \in \mathbb{R}^{T \times H \times W \times C}$ through three stages:

\textbf{Spatiotemporal Tokenization.} Videos are partitioned into non-overlapping tubelet patches of size $2 \times 16 \times 16$ (temporal $\times$ spatial). Each tubelet is flattened and linearly projected to $D=768$ dimensions:
\begin{equation}
\mathbf{z}_0 = [\mathbf{x}_{\text{cls}}; \mathbf{E}\mathbf{p}_1; \mathbf{E}\mathbf{p}_2; \ldots; \mathbf{E}\mathbf{p}_N] + \mathbf{E}_{\text{pos}}
\end{equation}
where $\mathbf{E} \in \mathbb{R}^{D \times (2 \cdot 16 \cdot 16 \cdot C)}$ is the pretrained patch embedding matrix, $\mathbf{x}_{\text{cls}}$ is a learned classification token prepended to the sequence, and $\mathbf{E}_{\text{pos}} \in \mathbb{R}^{(N+1) \times D}$ encodes spatiotemporal positional information.

\textbf{Transformer Encoder.} The token sequence passes through $L=12$ transformer layers. Each layer comprises Multi-Head Self-Attention (MSA, 12 heads) and Feed-Forward Network (FFN) with residual connections and Layer Normalization (LN):
\begin{align}
\mathbf{z}'_\ell &= \text{MSA}(\text{LN}(\mathbf{z}_{\ell-1})) + \mathbf{z}_{\ell-1}, \quad \ell = 1, \ldots, 12 \\
\mathbf{z}_\ell &= \text{FFN}(\text{LN}(\mathbf{z}'_\ell)) + \mathbf{z}'_\ell
\end{align}

The self-attention mechanism computes scaled dot-product attention:
\begin{equation}
\text{Attention}(\mathbf{Q}, \mathbf{K}, \mathbf{V}) = \text{softmax}\left(\frac{\mathbf{Q}\mathbf{K}^\top}{\sqrt{d_k}}\right)\mathbf{V}
\end{equation}
where $\mathbf{Q}, \mathbf{K}, \mathbf{V}$ are learned projections and $d_k = D/12 = 64$ is the dimension per head. This computes pairwise interactions between all tokens regardless of spatiotemporal distance, enabling global receptive fields that capture long-range dependencies across the entire video—essential for recognizing tackle dynamics spanning multiple phases (approach, contact, follow-through).

\textbf{Classification Head.} The final classification token $\mathbf{z}_{12}^{(0)}$ aggregates global video information through attention over all spatiotemporal tokens. This 768-dimensional representation is passed through a linear layer to produce 2 class logits, followed by softmax for class probabilities.

\subsection{Addressing Class Imbalance using focal loss}

we employ Focal Loss~\cite{lin2018focalloss} to address varying example difficulty. Focal Loss down-weights easy-to-classify examples and focuses learning on hard cases:
\begin{equation}
\mathcal{L}_{\text{FL}} = -\alpha_y (1 - p_y)^\gamma \log(p_y)
\end{equation}
where $p_y$ is the predicted probability for the true class $y \in \{0, 1\}$, $\gamma \geq 0$ is the focusing parameter, and $\alpha_y \in [0, 1]$ is the class-specific balancing weight.

\textbf{Focusing parameter $\gamma = 1.6$:} The modulating factor $(1-p_y)^\gamma$ approaches zero for well-classified examples ($p_y \to 1$), dramatically reducing their loss contribution. With $\gamma=1.6$, an example classified with 90\% confidence receives $(1-0.9)^{1.6} = 0.04$ weighting—a 25$\times$ reduction versus standard cross-entropy. Examples at 50\% confidence receive $(1-0.5)^{1.6} = 0.33$ weighting—a 3$\times$ reduction. This moderate focusing directs optimization toward challenging examples without completely ignoring easier cases.

\textbf{Balancing parameter $\alpha = 0.6$:} We assign $\alpha=0.6$ to risky tackles and $1-\alpha=0.4$ to safe tackles, providing near-neutral weighting that complements (rather than duplicates) the balanced sampling. This configuration prevents over-correction while ensuring the model attends appropriately to both classes.

\textbf{Synergy with transfer learning:} In fine-tuning scenarios, Focal Loss provides additional benefits beyond class balancing. The Kinetics-400 pretrained model already recognizes certain motion patterns similar to tackles, making some examples ``easy'' from initialization. Focal Loss automatically down-weights these well-transferred patterns while directing optimization toward tackle-specific features requiring task-specific adaptation. This prevents catastrophic forgetting of useful pretrained representations while refining decision boundaries for domain-specific nuances.

\section{Results and Discussion}

We explored augmentation configurations using a Taguchi $L_{18}$ orthogonal array, with each configuration evaluated via 5-fold cross-validation. In total, we conducted $18$ Taguchi-configured runs plus $2$ supplementary (original imbalance; duplication-only), each evaluated with 5-fold cross-validation, yielding $20 \times 5 =100$ trials.
 The first supplementary run uses the original imbalanced dataset with no augmentation or resampling; the second balances classes via duplication applied to the training set only, without augmentation.

Our primary evaluation metric is \emph{risky-class recall}, as correctly identifying risky tackle scenarios is critical for player safety, where missing a dangerous tackle (false negative) that could result in injury has far greater consequences than incorrectly flagging a safe tackle (false positive). We use the risky F1-score as a balanced secondary indicator.

A critical consideration is the instability introduced by data scarcity in the previously published validation set~\cite{Nafi2022RiskyTackle3D}. Their experiments used only 39 videos (26 safe, 13 risky). With just 13 risky examples, a single misclassification shifts risky-class recall by $\approx7.7$ percentage points ($1/13$), while one sample changes overall accuracy by $\approx2.6$ points ($1/39$). This volatility obscures meaningful performance differences between configurations. To obtain more stable and generalizable estimates, we expanded the validation set to 146 videos (95 safe, 51 risky). Under this regime, one risky sample alters recall by $\approx2.0$ points ($1/51$) and affects accuracy by $\approx0.68$ points ($1/146$), a roughly $3.8\times$ improvement in stability. This expanded dataset provides robust conclusions about augmentation effects and model performance.

\begin{figure}[t]
  \centering
  \includegraphics[width=\linewidth]{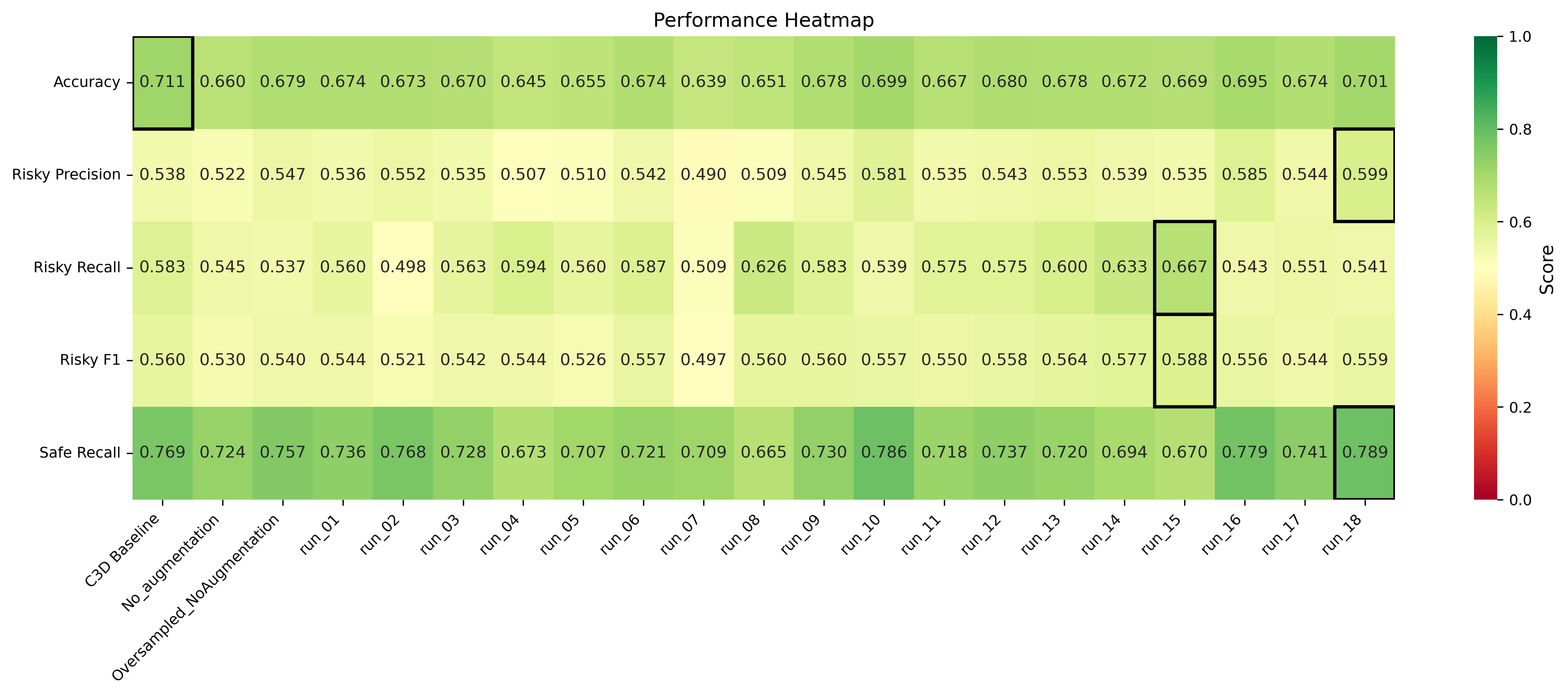}
  \caption{Performance heatmap showing mean scores across 5-fold cross-validation for all Taguchi configurations and supplementary runs. Rows correspond to evaluation metrics and columns to experimental runs. Cell values represent mean scores (0-1) computed using per-fold operating thresholds selected to maximize macro-F1. Black boxes highlight the best-performing configuration for each metric. $Run_{15}$ achieves optimal performance on both critical metrics: risky recall (0.67) and risky F1 (0.59).}
  \label{fig:performance_heatmap}
\end{figure}

\paragraph{Evaluation Metrics.}
We report fold-averaged results (mean over five folds) using confusion-matrix counts: True Positive (TP) as risky correctly detected, False Positive (FP) when safe is mislabeled as risky, True Negative (TN) when safe is correctly detected, and False Negative (FN) where risky is detected as safe. Metrics are computed as
$\text{Acc}=\frac{TP+TN}{TP+FP+TN+FN}$,
$\text{Prec}_{r}=\frac{TP}{TP+FP}$,
$\text{Rec}_{r}=\frac{TP}{TP+FN}$,
$\text{F1}_{r}=\frac{2TP}{2TP+FP+FN}$, and
$\text{Rec}_{s}=\frac{TN}{TN+FP}$.
Here Accuracy can be misleading under imbalance: predicting all samples as safe yields $\sim$65\% accuracy on our validation folds while detecting no risky tackles. We therefore treat risky recall as the primary endpoint, with risky F1 as a secondary metric to balance misses and false alarms.

\paragraph{Performance Overview.}
Figure~\ref{fig:performance_heatmap} summarizes all configurations and metrics, enabling direct comparison of augmentation strategies. $Run_{15}$ is best overall for the safety-critical objective, achieving the highest risky recall (0.67) and risky F1 (0.59), i.e., detecting roughly two-thirds of risky tackles with reasonable precision. Both supplementary baselines underperform (original imbalanced: risky recall 0.58; duplicated/oversampled: 0.54), indicating that systematic augmentation design is more effective than naive resampling.

Performance varies markedly across runs: several configurations reach high accuracy (0.70--0.71) but with reduced risky recall, illustrating the well-known trade-off between overall correctness and minority-class sensitivity under imbalance. $Run_{15}$ provides a favorable operating point, maintaining competitive accuracy (0.67) while prioritizing risky detection. $Run_{18}$ also performs well, particularly in safe recall and risky precision. Although the prior C3D baseline shows strong accuracy, accuracy alone is not decisive in this imbalanced, safety-critical setting.
\begin{figure}[t]
  \centering
  \includegraphics[width=\linewidth]{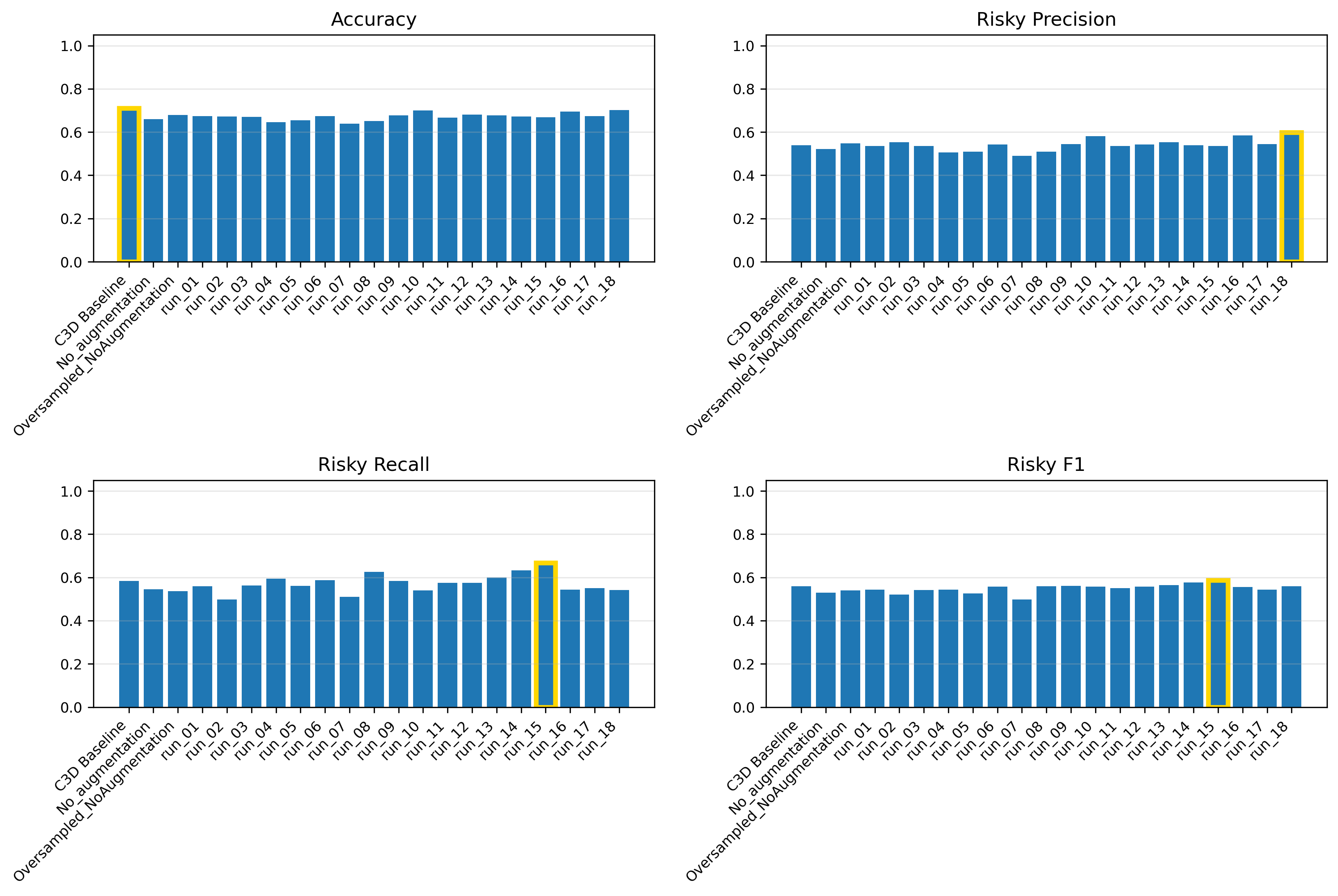}
  \caption{Detailed performance comparison across evaluation metrics for selected augmentation configurations. Run~15 (highlighted) demonstrates superior risky-class detection (recall = 0.67, F1 = 0.59) while maintaining balanced performance across metrics. The supplementary runs (original imbalanced and duplicated baseline) serve as reference points, illustrating the effectiveness of systematic augmentation design over naive class balancing.}
  \label{fig:metrics_bar}
\end{figure}

Figure~\ref{fig:metrics_bar} breaks down metrics for the strongest configurations and highlights these trade-offs: approaches that maximize accuracy often suppress risky recall, whereas methods tuned for risky detection may accept lower safe recall. $Run_{15}$ remains competitive across metrics while excelling at the primary requirement,identifying dangerous tackles.
\begin{figure*}[t]
  \centering
  \begin{subfigure}{0.48\textwidth}
    \centering
    \includegraphics[width=\linewidth]{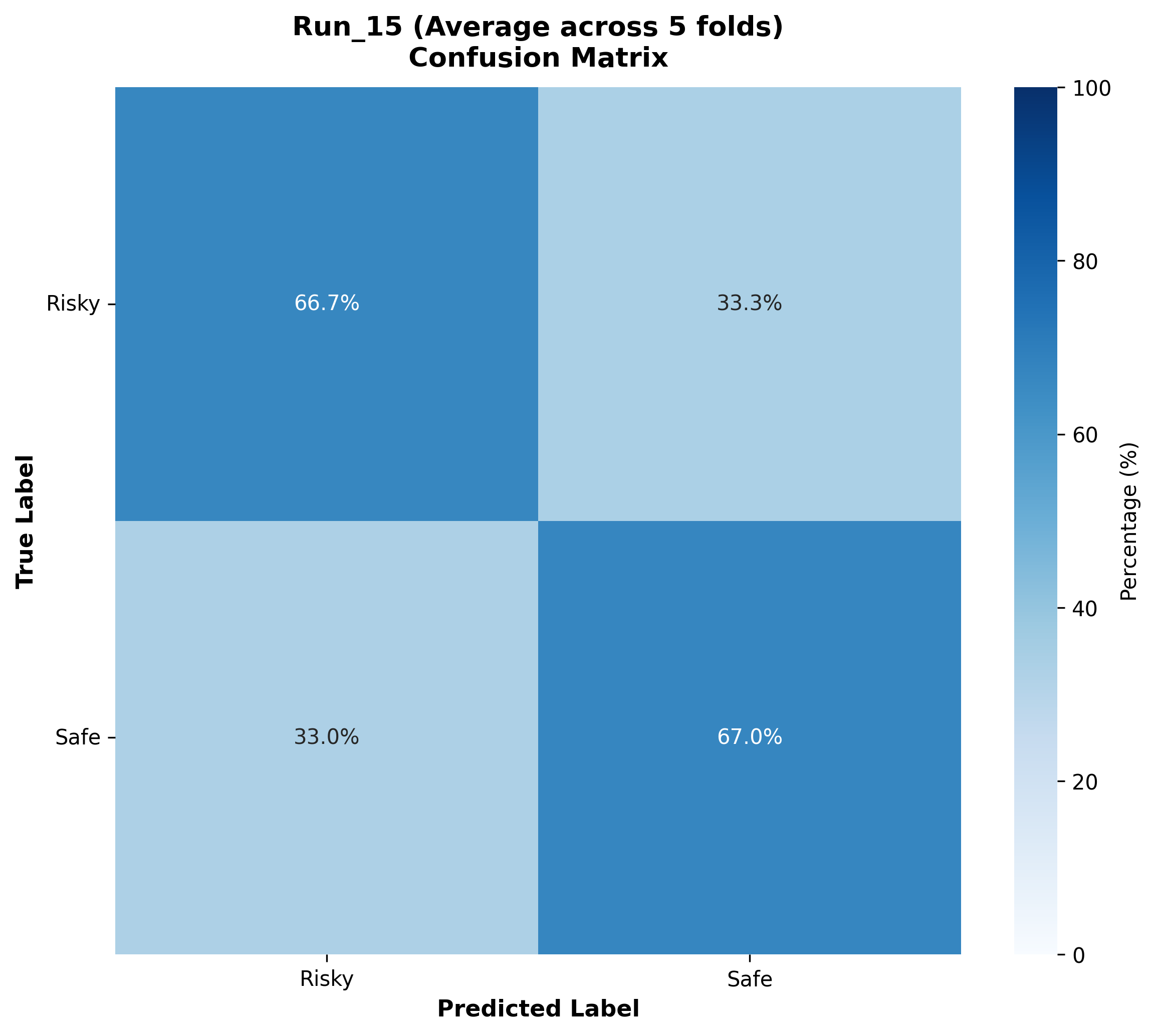}
    \caption{$Run_{15}$ (mean over 5 folds)}
    \label{fig:cm-run15}
  \end{subfigure}\hfill
  \begin{subfigure}{0.48\textwidth}
    \centering
    \includegraphics[width=\linewidth]{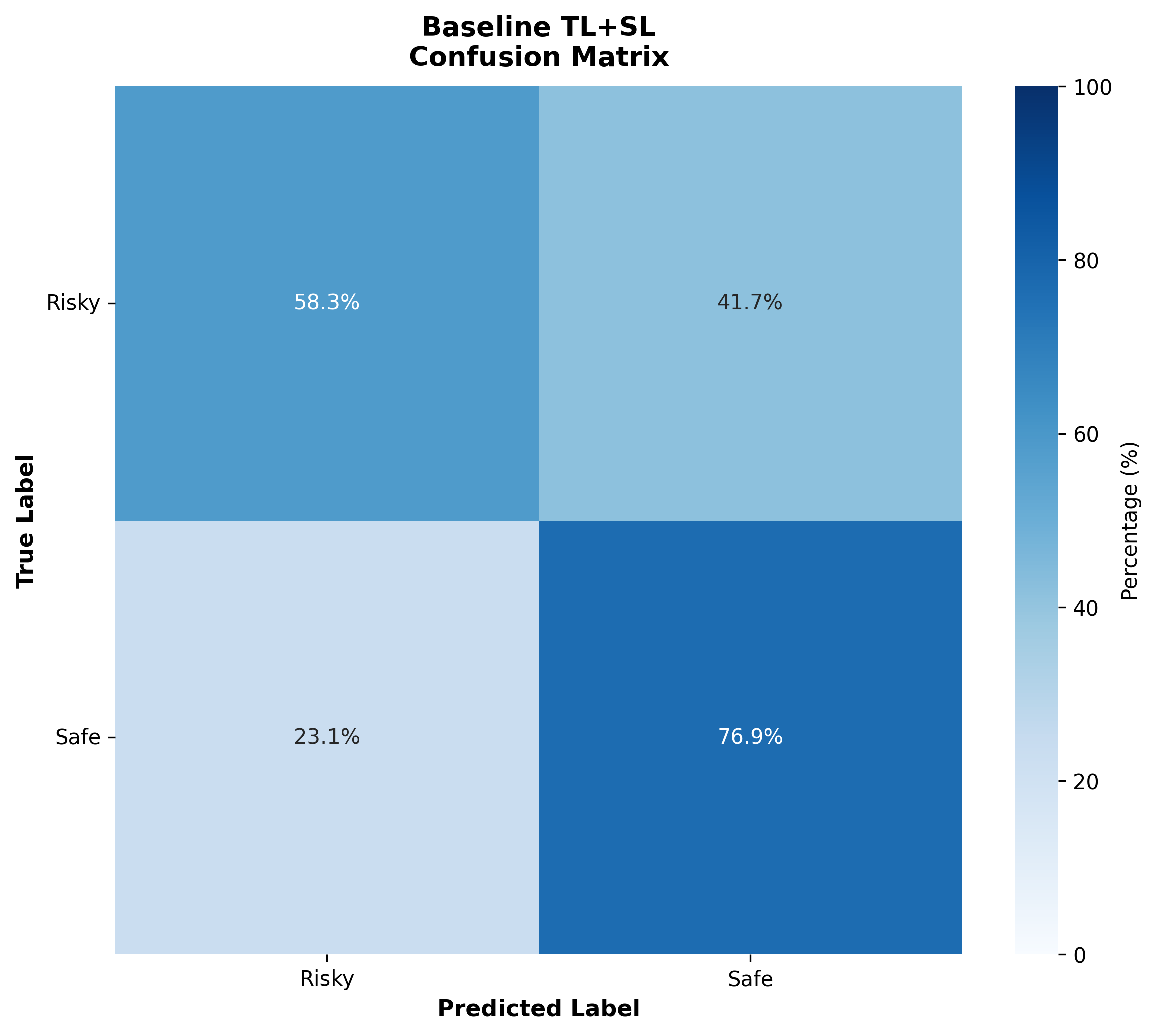}
    \caption{Baseline (original imbalanced)}
    \label{fig:cm-baseline}
  \end{subfigure}
  \caption{Confusion matrices normalized by class totals (percentages) for \textbf{(a)} $Run_{15}$ and \textbf{(b)}baseline configurations. $Run_{15}$ achieves $risky recall = 66.7\%$ and $safe recall = 67.0\%$, while the previous c3d baseline achieves $Risky Recall = 58.3\%$ and $Safe Recall = 76.9\%$. $Run_{15}$ improves risky-class detection by 8.4 percentage points at the cost of 9.9 points in safe recall, reflecting a more sensitive operating point that prioritizes identifying dangerous tackles, a critical requirement for player safety applications.}
  \label{fig:cms}
\end{figure*}
To examine error profiles, Fig.~\ref{fig:cms} compares percentage-normalized confusion matrices for $Run_{15}$ (mean over 5 folds) and the C3D baseline from prior work~\cite{Nafi2022RiskyTackle3D}. Because the baseline was evaluated on a smaller validation set (39 videos: 26 safe, 13 risky) than ours (146 videos: 95 safe, 51 risky) and we use 5-fold cross-validation, class-normalized matrices allow fair comparison. $Run_{15}$ attains 66.7\% risky recall and 67.0\% safe recall, while C3D attains 58.3\% risky recall and 76.9\% safe recall. Thus, $Run_{15}$ improves risky recall by 8.4 percentage points at the cost of a 9.9-point reduction in safe recall, consistent with our safety-critical priority of minimizing false negatives. The resulting increase in false positives is acceptable in a monitoring workflow where flagged plays can be reviewed, whereas missed dangerous tackles are far more costly.

\section{Conclusion}

This study addresses automated detection of risky tackles in American football, a safety-critical task complicated by class imbalance and limited training data. We systematically assessed how principled data augmentation influences video-based classification performance and make three contributions: (i) we expand the validation set from 39 to 146 videos, yielding markedly more stable estimates (approximately $3.8\times$ smaller per-sample metric step size); (ii) we explore augmentation strategies via a Taguchi $L_{18}$ orthogonal array, efficiently covering a broad design space within a constrained computational budget; and (iii) we identify a best-performing augmentation configuration within this design that improves detection of dangerous tackles under 5\,-fold cross-validation.

Fine-tuning a pretrained Video Vision Transformer~\cite{arnab2021vivit} with focal loss for imbalance mitigation, our top configuration (\texttt{$Run_{15}$}) achieves \textbf{66.7\%} risky recall and \textbf{59.0\%} risky F1, a \textbf{+8.4} percentage point gain in risky recall over the baseline C3D (58.3\%). This recipe combines photometric perturbations (Gaussian noise and brightness reduction) while omitting spatial rotations and flips. The increased sensitivity to dangerous tackles is accompanied by a 9.9 point decrease in safe recall, a deliberate trade-off in a safety-monitoring context where false negatives are more consequential than false positives that can be triaged by human review.

Our results suggest that \emph{photometric} variability (e.g., noise/brightness adjustments) is especially beneficial for this task, whereas aggressive \emph{geometric} perturbations can be less helpful. A plausible explanation is that preserving spatial body configuration and relative positioning features that correlate with tackle mechanics is important for safety assessment. Overall, carefully designed augmentation pipelines can mitigate data scarcity and imbalance in video activity recognition without architectural changes or additional labels.

\paragraph{Limitations and future work.}
Although the expanded validation set improves stability relative to prior work, our evaluation reflects a single team and practice context; generalization to different teams, camera viewpoints, lighting, and competitive settings remains to be validated. The optimal augmentation intensities identified here may also be dataset-dependent. Promising directions include: (i) systematic tuning of augmentation intensity schedules (e.g., brightness ranges, noise magnitudes); (ii) evaluating modern video transformers such as Swin~\cite{SWINliu2021} and MViTv2~\cite{mvitv2}; (iii) exploring cost-sensitive objectives beyond focal loss, calibration strategies, and ensembles; and (iv) advancing toward real time deployment with temporal action localization to flag precise contact moments. Progress along these lines can support coaches and medical staff in earlier identification of high-risk plays and inform safety-oriented policy and training interventions.

%
%
%
%
  \bibliographystyle{elsarticle-num} 
  \bibliography{references}
\end{document}